\pdfoutput=1
\documentclass[11pt]{article}
\usepackage{multirow}

\usepackage[preprint]{acl}
\usepackage{times}
\usepackage{latexsym}
\usepackage{tikz}
\usepackage{amsmath}
\usepackage{amssymb}  % 提供额外的数学符号
\usepackage{authblk}  % 用于处理作者和机构的排版
\usepackage[T1]{fontenc}
\usepackage{booktabs}
\usepackage[utf8]{inputenc}

\usepackage{microtype}

\usepackage{inconsolata}

\usepackage{graphicx}

\title{Large Language Models Badly Generalize across Option Length, Problem Types, and Irrelevant Noun Replacements}

\author{
\vspace{-0.4cm} 
  Guangxiang Zhao*$^1$, Saier Hu*$^2$, Xiaoqi Jian*$^2$, Jinzhu Wu$^2$, Yuhan Wu$^3$, \\  \vspace{0.2cm} 
  Lin Sun$^2$\textsuperscript{\dag}, Xiangzheng Zhang$^2$\textsuperscript{\dag} \\\vspace{0.2cm} 
  $^1$Qiyuan Tech, $^2$360Zhinao, $^3$Peking University \\
  \texttt{\{zhaoguangxiang, yuhan.wu\}@pku.edu.cn}, \\
  \texttt{\{husaier, jianxiaoqi, wujinzhu, sunlin1, zhangxiangzheng\}@360.cn}
  \thanks{Equal Contribution}
  \thanks{Corresponding Authors}
}

\begin{document}
% 我在写一篇论文，准备投稿深度学习会议ICLR， 请你用英文建议几个标题，建议第一页的图描绘什么，写一段abstract，一段introduction, 和给出论文框架， 并给我绘制第一页图片的细节，可能的话，给相关代码。我目前感觉研究是关于大模型评测的缺陷，发现评测集里的题目稍微改下格式，就会导致大模型的评测准确率下降很多，目前的3个发现如下：1. 将评测集里的选择题改成填空题或者判断题，大模型准确率都会有很大变化，其中三类题型都做对的不足一半。 2. 将选择题选项的长度改变，比如只把三个错误选项的一个选项改长，模型的准确率大幅下降, 并且发现模型倾向于选那个错误的长选项。 3. 把评测集题目中对答案没影响的无关名词进行替换，发现模型预测准确率大幅下降。
% 可以给出abstract和introduction的思维导图吗
%最近有人研究过大模型对评测不鲁棒，比如发现大模型对选项的编号形式不鲁棒，对few-shot中样本的答案分布不鲁棒等，我该如何修改abstract和introduction

% 我在写一篇论文，准备投稿深度学习会议ICLR， 请你用英文建议几个标题，建议第一页的图描绘什么，写一段abstract，一段introduction, 和给出论文框架。我目前感觉研究是关于大模型评测的缺陷，发现评测集里的题目稍微改下格式，就会导致大模型的评测准确率下降很多，目前的3个发现如下：1. 将评测集里的选择题改成填空题或者判断题，大模型准确率都会有很大变化，其中三类题型都做对的不足一半。 2. 将选择题选项的长度改变，比如只把三个错误选项的一个选项改长，模型的准确率大幅下降, 并且发现模型倾向于选那个错误的长选项。 3. 把评测集题目中对答案没影响的无关名词进行替换，发现模型预测准确率大幅下降。

% Please write some attractive titles
% Should the title reflect the three mentioned aspects?
% 你觉得这三方面是否都算是格式问题
% 标题的不同会对论文写作产生怎样的影响
% 有相关论文已经说了点大模型的不鲁棒性，比如把选项的abcd改成数字会降低结果，这些相关工作将对我的论文标题和写作产生怎样的影响
\maketitle
\begin{abstract}
% we investigate the fragility of Large Language Models (LLMs) in generalizing to novel inputs, specifically focusing on minor perturbations in well-established benchmarks (e.g., slight changes in problem type or distractor length). 
In this paper, we propose a “Generalization Stress Test” to assess  Large Language Models' (LLMs) generalization ability under slight and controlled perturbations, including option length, problem types, and irrelevant noun replacements. 
We achieve novel and significant findings that, despite high benchmark scores, LLMs exhibit severe accuracy drops and unexpected biases (e.g., preference for longer distractors) when faced with these minor but content-preserving modifications. For example, Qwen 2.5 1.5B's MMLU score rises from 60 to 89 and drops from 89 to 36 when option lengths are changed without altering the question. Even GPT4o experiences a 25-point accuracy loss when problem types are changed, with a 6-point drop across all three modification categories.
 These analyses suggest that LLMs rely heavily on superficial cues rather than forming robust, abstract representations that generalize across formats, lexical variations, and irrelevant content shifts. Code can be found in:~\url{https://github.com/Qihoo360/LLMs-Generalization-Test}.
 
 % This work proposes a “Generalization Stress Test” to assess performance shifts under controlled perturbations and calls for developing more reliable evaluation methodologies to capture LLM generalization abilities better.

\end{abstract}

\section{Introduction}\label{sec_intro}
Large Language Models (LLMs) have achieved near-human performance across a variety of natural language processing (NLP) benchmarks, from elementary tests \citep{gsm} to university-level challenges \citep{mmlu}. This success has spurred claims that LLMs are approaching human-like generalization capabilities \citep{gpt4, bubeck2023sparksartificialgeneralintelligence, jones2024peopledistinguishgpt4human}. However, it remains unclear whether their high benchmark scores reflect genuine generalization or if LLMs are simply exploiting superficial cues that fail under slight perturbations.

While LLMs perform well in established benchmarks, concerns have been raised about the validity of these evaluations~\cite{chen2023robustgpt35predecessorscomprehensive, ye2023comprehensivecapabilityanalysisgpt3}. Data contamination, where models unintentionally learn from benchmark data included in their training, can inflate performance estimates \citep{gpt3, data_conta_bench, data_conta_much, zhou2023dontmakellmevaluation}. These issues suggest that existing benchmarks have exposed patterns and may not truly assess generalization.

Recent work has focused on uncovering the actual limits of LLM generalization.  One direction involves the development of dynamic evaluation methods that modify the evaluation process on the fly \citep{dyval, data_conta_dynamic_eval}, or extend the modality~\cite{wang2025reliablevlmfinegrainedbenchmark}. Another approach emphasizes creating more challenging or adversarial test sets that push models beyond their current capabilities, such as MMLU-Pro \citep{mmlu_pro} and GSM-Plus \citep{gsm_plus}. A third line of inquiry involves introducing subtle modifications to benchmark datasets to test LLM robustness, such as altering the order of multiple-choice options or changing the format of questions \citep{robust_multichoice, li-etal-2024-multiple, robust_change_answer_order, roubst_benchmarks_target, hong2025qualbench}. While these approaches have contributed to a better understanding of LLM performance, they either totally change the original problems, increase the complexity of the evaluation, or focus on relatively limited formatting changes like option ID adjustments.

\begin{figure*}[t]
  \includegraphics[width=1.0\linewidth]{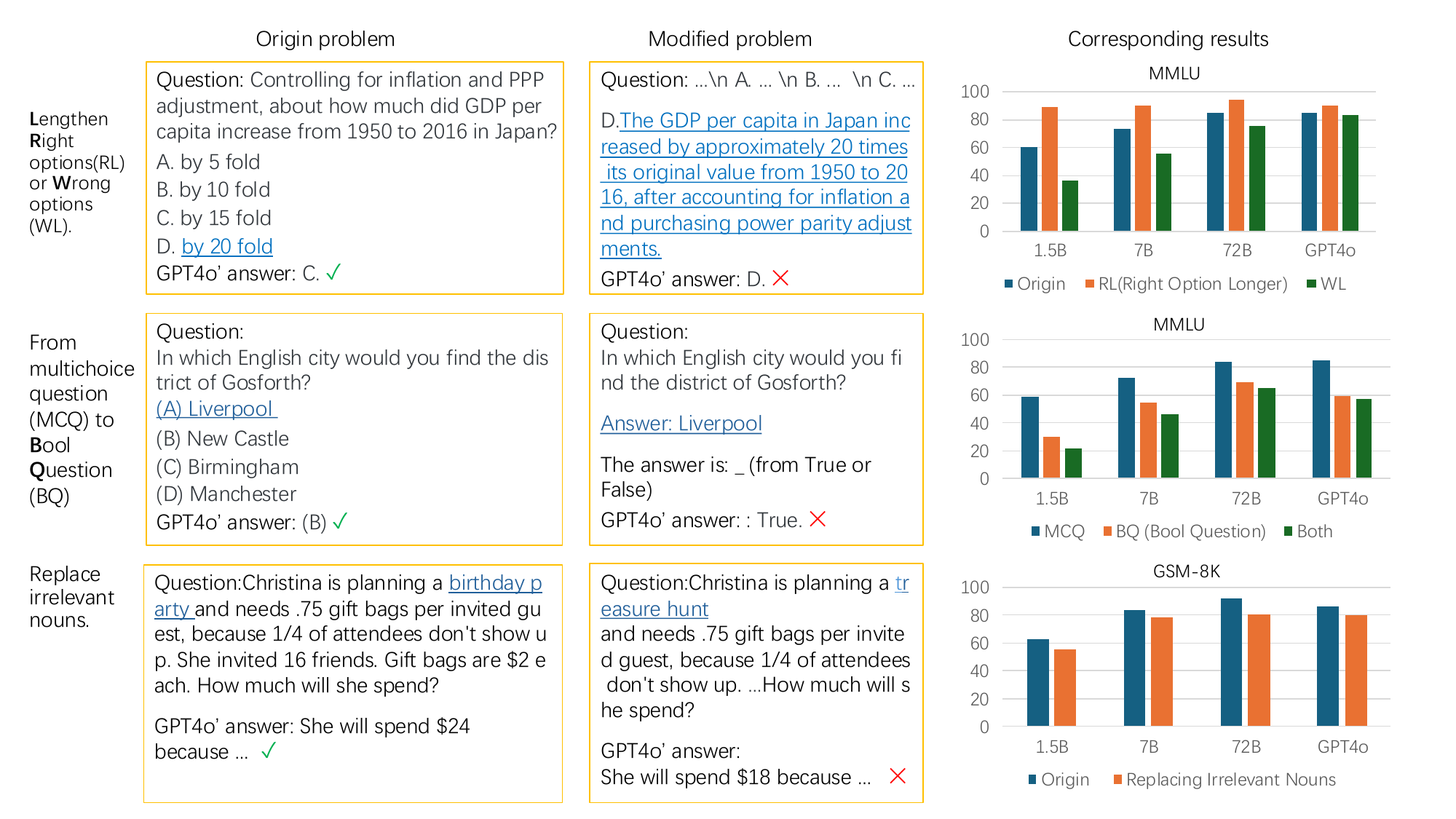} 
  \caption {Generalization stress tests and summarized results. LLMs do not generalize well across various option lengths, problem types, and noun replacements. Tested models are Qwen2.5 1.5B, 7B, 72B, and GPT4o. }
  \label{figureintro}
\end{figure*}

We find serious biases of recent SoTa LLMs to common patterns by introducing an evaluation framework, \textbf{Generalization Stress Tests}, which examines LLMs under three types of minor, content-preserving perturbations: 
\begin{itemize}
    \item Altering option length (e.g., increasing the length of distractors or correct options without changing their semantic content).
    \item Changing problem types (e.g., converting multiple-choice questions to boolean questions).
    \item Replacing irrelevant nouns (e.g., substituting semantically irrelevant nouns in prompts).
\end{itemize}

% As shown in Figure  \ref{figureintro}, these minor changes lead to significant performance degradation, revealing that LLMs rely on superficial cues, such as question format and option length, rather than robust, abstract representations. Our results highlight the limitations of current LLM evaluation methods and call for the development of more robust benchmarks that better capture the true generalization capabilities of these models.
As shown in Figure \ref{figureintro}, these simple modifications, surprisingly, lead to substantial performance degradation\footnote{We test GSM-8K for noun replacement, as some MMLU cases lack irrelevant nouns.}. We observe that LLMs struggle to generalize across varying option lengths, problem types, and noun replacements. For example, Qwen 2.5 1.5B's MMLU score drops from 89 to 36 when option lengths are changed without altering the question. Even GPT4o experiences a 25-point accuracy loss when question types are changed, with a 6-point drop across all three categories. These findings reveal a critical limitation: LLMs are biased to specific irrelevant patterns and fail to replicate the human-like ability to ignore irrelevant format details.

\section{Methods: Generalization Stress Tests}
We conduct generalization stress tests by applying minor modifications to the original benchmark, focusing on variations in option length, scoring type, and the replacement of irrelevant nouns.

We investigate typical tasks for LLMs that include multiple-choice questions (MCQ) and open-ended question answering (Open-ended QA). 
% An MCQ task involves a question with several answer choices, with one correct option and the others incorrect (e.g., MMLU\cite{mmlu}). In contrast, open-ended QA requires the model to provide a free-form answer to a question without candidate options (e.g., GSM-8K\cite{gsm}).

% Evaluation benchmarks for large models, such as GSM8K, MATH\cite{math}, and OpenAI HumanEval\cite{humaneval}, are in this form.  However, previous work on examining robustness rarely tests this type of benchmark.

% Some research has found vulnerabilities in this type of question\cite{robust_multichoice}, while other studies have found it to be more robust compared to other types\cite{zhang2024multiplechoicequestionsefficientrobust}. 

\subsection{Alter Option Length to Analyze LLMs' Length Bias}
\begin{figure}[h!]
    \centering
    \begin{tikzpicture}
        % Multiple-choice question box
        \node[draw, rectangle, rounded corners, minimum width=7cm, minimum height=3cm, fill=blue!10] (mcq) at (0, 3.5) {
            \begin{minipage}{7cm}
                \textbf{Make the right option longer (RL):} \\
                Question: What is the capital of France? \\
                A) Berlin \\ B) Madrid\\ C) Paris, a city renowned for its art, fashion, and cuisine.   \\D) Rome 
            \end{minipage}
        };
        % Open-ended question box
        % \node[draw, rectangle, rounded corners, minimum width=7cm, minimum height=3cm, fill=green!10] (oeq) at (0, -0.7) {
        %     \begin{minipage}{7cm}
        %         \textbf{Make all wrong option longer (WL-ALL):} \\
        %         Question: What is the capital of France? \\
        %         A) Berlin, known for its vibrant culture and historical landmarks. \\ B) Madrid, famous for its lively nightlife and beautiful architecture.\\ C) Paris \\D) Rome, celebrated for its ancient ruins and rich history. 
        %     \end{minipage}
        % };
        \node[draw, rectangle, rounded corners, minimum width=7cm, minimum height=3cm, fill=pink!10] (oeq) at (0, -0.3) {
            \begin{minipage}{7cm}
                \textbf{Make one wrong option longer (WL):} \\
                Question: What is the capital of France? \\
                A) Berlin, known for its vibrant culture and historical landmarks. \\ B) Madrid\\ C) Paris  \\D) Rome
            \end{minipage}
                };
        % Labels
        % \node at (0, 4) {\textbf{Illustration of Question Types}};
    \end{tikzpicture}
     \caption{An illustration of altering option length. The ground truth of this question is C) Paris. }
    \label{fig:question_lenghten_types}
\end{figure}
% In RL, we make the right option (C) longer. We make all wrong options (A, B, D) longer in WL-ALL. In WL, we randomly make one wrong option (A) longer.
To analyze whether LLMs are generalized across option length or whether LLMs are biased toward long options in MCQ.  We first make all options in a problem longer by asking GPT4o\footnote{We use its API version provided by Microsoft Azure.} to make the options longer without including information that could help answer the question. Refer to Appendix~\ref{sec:appendix-length} for generation details.

As illustrated in Figure \ref{fig:question_lenghten_types}, we then design the following two types of lengthening problems: a)Make one wrong option longer (WL), b)Make the right options longer (RL).
 % b) Make all wrong options longer (WL-ALL).

\noindent\textbf{Length Control:} To assess the impact of option length on LLM generalization, we control the length of the lengthened options in the WL condition. Specifically, we ask GPT4o to generate options of varying lengths: (a) < 10 tokens, (b) 10 to 20 tokens, and (c) > 20 tokens.

\noindent\textbf{Paraphrase Verification:} We also enlist human experts to verify whether the paraphrased options do not introduce unintended biases or hints. Details can be found in the Appendix~\ref{sec:appendix-length}.
%这里要给表格说明人工和4o检查给出的百分比，可以写在附录

\subsection{Change Problem Type to Fairly Analyze LLMs' Scoring Bias} 
%这里太突兀，不好和故事串起来，且方法太简单
\begin{figure}[h!]
    \centering
    \begin{tikzpicture}
        % Multiple-choice question box
        \node[draw, rectangle, rounded corners, minimum width=7cm, minimum height=2cm, fill=cyan!10] (mcq) at (0, 2.5) {
            \begin{minipage}{7cm}
                \textbf{Cloze:} \\
                Question: What is the capital of France? \\
                Answer: \_ (\textit{Selected from \textbf{whole vocabulary}})
            \end{minipage}
        };
        % Open-ended question box
        \node[draw, rectangle, rounded corners, minimum width=7cm, minimum height=2.5cm, fill=magenta!10] (oeq) at (0, -0.9) {
            \begin{minipage}{7cm}
                \textbf{Bool questions:} \\ 
                1. Question: What is the capital of France? \\
                Answer: Paris \\
                The answer is \_(\textit{Selected from True/False}) \\
                2. Question: What is the capital of France? \\
                Answer: Berlin \\
                The answer is \_(\textit{Selected from True/False}) \\
                \textit{Require to judge both two propositions correctly.}
            \end{minipage}
        };
        % Labels
        % \node at (0, 4) {\textbf{Illustration of Question Types}};
    \end{tikzpicture}
     \caption{An illustration of changing the scoring type from MCQ to bool questions.}
    \label{fig:bq}
\end{figure}
Previous work found LLMs do not generalize to different option IDs in MCQ \cite{robust_multichoice} and tried to solve this by changing the task to cloze~\cite{roubst_benchmarks_target}. However, the cloze task reduces the expected value of selecting the correct answer. Therefore, we propose changing the multiple-choice questions to Boolean questions, requiring two judgments to be accurate, so that the difficulty of the questions is as similar as possible to that of multiple-choice questions. 

As illustrated in Figure \ref{fig:bq}, we derive one true proposition that concludes with the right option and one false proposition that is a randomly selected wrong option. 
% 这里可能需要多做一些分析
\subsection{Replace Irrelevant Nouns to Analyze Bias towards Irrelevant Content}
\begin{figure}[h!]
    \centering
    \begin{tikzpicture}
        % Multiple-choice question box
        \node[draw, rectangle, rounded corners, minimum width=7cm, minimum height=3cm, fill=blue!10] (mcq) at (0, 2.7) {
            \begin{minipage}{7cm}
                \textbf{Problem with irrelevant noun:} \\
                Question: \textbf{John} lives in France; what is his country's capital? \\
                A) Berlin \\ B) Madrid\\ C) Paris  \\D) Rome \\ Answer: C
            \end{minipage}
        };
        % Open-ended question box
        \node[draw, rectangle, rounded corners, minimum width=7cm, minimum height=1cm, fill=green!10] (oeq) at (0, -1.7) {
            \begin{minipage}{7cm}
                \textbf{Problem after modifying the irrelevant noun:} \\
                Question: \textbf{Mike} lives in France; what is his country's capital? \\
                A) Berlin \\ B) Madrid\\ C) Paris  \\D) Rome \\ Answer: C

            \end{minipage}
        };
        % Labels
        % \node at (0, 4) {\textbf{Illustration of Question Types}};
    \end{tikzpicture}
    \caption{An illustration of replacing irrelevant nouns.}
    \label{fig:irrelevent}
\end{figure}
In open-ended QA like those in GSM8K~\cite{gsm}, the questions may contain nouns that are unrelated to the answers. In this subsection, we explore the impact of changes to these unrelated nouns on the decision-making of large models. As shown in Figure \ref{fig:irrelevent}, we replaced nouns in the questions, such as names of people and animals, ensuring that these replacements do not alter human decision-making. Details are in Appendix~\ref{sec:appendix-noun}.

\noindent\textbf{Semantic relevance control} Additionally, regarding noun replacements, we also examined the impact of the semantic proximity of the replacements. We conducted experiments in this area by instructing GPT-4o mini to perform replacements with varying degrees of semantic similarity. 

\section{Experiments}~\label{sec_experiment}
We perform evaluations on harness framework \cite{eval-harness} and adopt its default setting. We evaluate models of Llama3.1 series \cite{llama31}, Qwen2.5 series \cite{yang2024qwen2}, and GPT4o. Llama3.1, and Qwen2.5 are the most powerful small models, while GPT4o is the most powerful LLM. We evaluate LLMs on MMLU~\cite{mmlu}, ARC-Challenge~\cite{arc}, and GSM8k~\cite{gsm}. The first two are MCQ benchmarks, and the last consists of open-ended QAs. Refer to Appendix~\ref{sec:appendix-exp} for detailed experimental setups.

\subsection{Results of Altering Option Length}
\begin{table}
  \centering
  \small
  \begin{tabular}{lllll}
    \hline
    \textbf{Benchmark} & \textbf{Model}& \textbf{Origin}  & \textbf{RL} & \textbf{WL} \\
    \hline
    \multirow{8}{*}{MMLU} & Qwen2.5 1.5B & \textbf{60.3} & \textbf{89.0} & \textbf{36.3}\\
                           & Qwen2.5 7B & 73.7 &90.1  &55.6 \\
                           & Qwen2.5 72B & 85.4 & 94.1 & 75.6 \\
                           % &Qwen2 1.5B &55.8 &90.0 &27.3\\
                           % & Qwen2 7B & 70.6 & 87.8 & 57.5 \\
                           % & Qwen2 72B & 83.8 & 94.1 & 74.1 \\
                           & LLaMa3.1 8B & 65.5 & 85.6 & 53.6 \\
                           & LLaMa3.1 70B & 78.8 & 93.6 & 70.6 \\
                           & GPT4o mini & 76.5 & 87.2 & 70.6 \\
                           & GPT4o & 85.2 & 89.7 & 83.3 \\
    \hline
    % \multirow{8}{*}{ARC-C} & Qwen2 1.5B & 69.0 & 88.3 & 53.8 \\
    %                        & Qwen2 7B & 87.2 & 93.3 & 83.8 \\
    %                        & Qwen2 72B & 95.7 & 97.3 & 93.1 \\
    \multirow{8}{*}{ARC-C} & Qwen2.5 1.5B & 77.3 & 88.9 & 68.1 \\
                           & Qwen2.5 7B & 90.0 & 94.3 & 84.0 \\
                           & Qwen2.5 72B & 95.8 & 97.2 & 94.4 \\
                           & LLaMa3.1 8B & 78.1 & 85.2 & 74.7 \\
                           & LLaMa3.1 70B & 91.8 & 96.3 & 90.8 \\
                           & GPT4o mini & 91.8 & 95.1 & 91.4 \\
                           & GPT4o & 96.5 & 97.1 & 95.5 \\
    \hline
  \end{tabular}
  \caption{\label{tab:len_res}
    Performance on altering option length. RL refers to lengthening the right option; WL refers to lengthening the wrong option. The values are percentages.
  }
\end{table}

\noindent \textbf{LLMs struggle to generalize across option length:} From Table \ref{tab:len_res}, it is evident that across all LLMs, from 1.5B to GPT4o, scores increase significantly when the length of the correct option is extended and decrease significantly when we make an incorrect option longer. Smaller models generalize even worse.
In Appendix~\ref{sec:al}, we introduce another setting of making all options longer, in which our finding that LLMs are biased towards the longer option persists. 
% For instance, on MMLU, when the correct option length is increased, the score for Qwen2 1.5B improves from 55.8 to 90.0. However, when we make a wrong option longer, the score drops sharply from 55.8 to 27.3, less than half. Even for the two 70B models, there remains a gap of over 20 points between RL and WL, with GPT4o showing a 6-point difference. 

\noindent \textbf{Length matters, especially when we lengthen the right option.} As shown in Table \ref{tab:vary_length}, changing the length can result in a difference of more than 10 points in the RL setting.
\begin{table}
\centering
\small
\begin{tabular}{|c|c|c|c|}
\hline
\textbf{Settings} & \textbf{<10} & \textbf{10 to 20} & \textbf{>20} \\ \hline
\textbf{Origin} & \multicolumn{3}{c|}{65.5\%} \\ \hline
\textbf{RL} & 70.0\% & 75.3\% & 84.0\% \\ 
\textbf{WL} & 64.5\% & 60.7\% & 61.6\% \\ 
\hline
\end{tabular}
\caption{The performance of LLaMa3.1 8B on MMLU changes when gradually altering the length of correct and incorrect options.}
\label{tab:vary_length}
\end{table}

Another intriguing finding is that LLMs tend to select the right option if we make all incorrect options longer, refer to Appendix~\ref{sec:wl_all}.

% \begin{table}
%   \centering
%   % \small
%   \begin{tabular}{llll}
%     % \hline
%     \multicolumn{4}{c}{\textbf{MMLU}} \\
%     \hline
%     \textbf{Model} &\textbf{Origin}           & \textbf{RL} & \textbf{WL} \\
%     \hline
%     Qwen2 1.5B &55.8 &90.0 &27.3\\
%     % Gemma2 2B &37.0 &50.5 &28.7\\
%     Qwen2 7B&70.6 &87.8 &57.5\\
%     LLaMa3.1 8B& 65.5 &85.6 &99.9\\
%     % Gemma2 9B&48.7 &55.7 &43.5\\
%     LLaMa3.1 70B&78.8 &93.6 &70.6\\
%     Qwen2 72B&83.8 &94.1 &74.1\\
%     GPT4o mini&76.5 &87.2 &70.6\\
%     GPT4o&85.2 &89.7 &83.3\\
%     \hline
%     \\
%     \multicolumn{4}{c}{\textbf{ARC-Challenge}} \\
%     \hline
%     \textbf{Model} &\textbf{Origin}           & \textbf{RL} & \textbf{WL} \\
%     \hline
%     Qwen2 1.5B & 69.0 & 88.3 & 53.8 \\
%    % & Gemma2 2B & 40.6 & 47.2 & 39.6 \\
%     Qwen2 7B & 87.2 & 93.3 & 83.8 \\
%     LLaMa3.1 8B & 78.1 & 85.2 & 74.7 \\
%    % & Gemma2 9B & 53.0 & 54.4 & 54.8 \\
%     LLaMa3.1 70B & 91.8 & 96.3 & 90.8 \\
%     Qwen2 72B & 95.7 & 97.3 & 93.1 \\
%     GPT4o mini & 91.8 & 95.1 & 91.4 \\
%     GPT4o & 96.5 & 97.1 & 95.5 \\ \hline
%   \end{tabular}
%   \caption{\label{fig:len_res}
%     Performance on altering option length. RL refers to making the right option longer; WL refers to making a random long option longer.
%   }
% \end{table}

\subsection{Results of Altering Scoring Type}
\noindent\textbf{LLMs do not have invariant knowledge that can generalize across scoring types.} As in Table \ref{fig:bq_res}, all models tend to score lower when the benchmarks are changed from the original format to boolean questions. Qwen2.5 1.5B and Llama3.1 8B score only half the points in the MMLU's "both" setting. Smaller models generalize worse.\footnote{The ``MCQ'' setting is equal to ``Origin'' setting in Table \ref{tab:len_res}, the results are slightly different since we removed the instruction “Output the answer directly” to accommodate the BQ setting.}
% \noindent\textbf{LLMs do not have invariant knowledge that can generalize across scoring types.}

% \begin{table}
%   \centering
%   % \small
%   \begin{tabular}{llll}
%     % \hline
%     \multicolumn{4}{c}{\textbf{MMLU}} \\
%     \hline
%     \textbf{Model} &\textbf{MCQ}           & \textbf{BQ} & \textbf{Both} \\
%     \hline
%     Qwen2 1.5B &55.8 &90.0 &27.3\\
%     % Gemma2 2B &37.0 &50.5 &28.7\\
%     Qwen2 7B&70.6 &87.8 &57.5\\
%     LLaMa3.1 8B& 65.5 &85.6 &99.9\\
%     % Gemma2 9B&48.7 &55.7 &43.5\\
%     LLaMa3.1 70B&78.8 &93.6 &70.6\\
%     Qwen2 72B&83.8 &94.1 &74.1\\
%     GPT4o mini&76.5 &87.2 &70.6\\
%     GPT4o&85.2 &89.7 &83.3\\
%     \hline
%     \\
%     \multicolumn{4}{c}{\textbf{ARC-Challenge}} \\
%     \hline
%     \textbf{Model} &\textbf{MCQ}           & \textbf{BQ} & \textbf{Both} \\
%     \hline
%     Qwen2 1.5B & 69.0 & 88.3 & 53.8 \\
%    % & Gemma2 2B & 40.6 & 47.2 & 39.6 \\
%     Qwen2 7B & 87.2 & 93.3 & 83.8 \\
%     LLaMa3.1 8B & 78.1 & 85.2 & 74.7 \\
%    % & Gemma2 9B & 53.0 & 54.4 & 54.8 \\
%     LLaMa3.1 70B & 91.8 & 96.3 & 90.8 \\
%     Qwen2 72B & 95.7 & 97.3 & 93.1 \\
%     GPT4o mini & 91.8 & 95.1 & 91.4 \\
%     GPT4o & 96.5 & 97.1 & 95.5 \\ \hline
%   \end{tabular}
%   \caption{\label{fig:bq_res}
%     Performance on changing problem type from multi-choice question (MCQ) to bool questions (BQ).
%   }
% \end{table}

\begin{table}
  \centering
  \small
  \begin{tabular}{lllll}
    \hline
    \textbf{Benchmark} & \textbf{Model}& \textbf{MCQ}  & \textbf{BQ} & \textbf{Both} \\
    \hline
    % \multirow{8}{*}{MMLU} & Qwen2 1.5B & 53.5 & 32.6 & 24.4 \\
    %                        & Qwen2 7B & 69.8 & 52.4 & 44.1 \\
    %                        & Qwen2 72B & 83.8 & 68.7 & 64.2 \\
        \multirow{8}{*}{MMLU} & Qwen2.5 1.5B & 58.8 & 30.3 & 22.1 \\
                           & Qwen2.5 7B & 72.4 & 54.7 & 46.7 \\
                           & Qwen2.5 72B & 84.0 & 69.1 & 65.0 \\
                           & LLaMa3.1 8B & 64.6 & 40.6 & 32.6 \\
                           & LLaMa3.1 70B & 78.4 & 63.5 & 56.7 \\
                           & GPT4o mini & 75.1 & 54.5 & 49.2 \\
                           & GPT4o & 84.7 & 59.5 & 56.8 \\
    \hline
    % \multirow{8}{*}{ARC-C} & Qwen2 1.5B & 68.6 & 35.4 & 32.3 \\
    %                        & Qwen2 7B & 87.1 & 71.2 & 67.5 \\
    %                        & Qwen2 72B & 95.7 & 85.8 & 84.4 \\
    \multirow{8}{*}{ARC-C} & Qwen2.5 1.5B & 74.0 & 40.4 & 35.2 \\
                           & Qwen2.5 7B & 89.5 & 69.4 & 66.4 \\
                           & Qwen2.5 72B & 95.0 & 85.8 & 84.4 \\
                           & LLaMa3.1 8B & 77.4 & 53.6 & 47.1 \\
                           & LLaMa3.1 70B & 92.1 & 82.7 & 79.2 \\
                           
                           & GPT4o mini & 90.6 &79.7  & 76.6 \\
                           & GPT4o & 96.2 & 79.6 & 76.2 \\
    \hline
  \end{tabular}
  \caption{\label{fig:bq_res}
    Performance on changing problem type from multi-choice question (MCQ) to bool questions (BQ). The values are percentages. ``Both'' means the percentages of examples whose MCQ and BQ are both true.
  }
\end{table}

\subsection{Results of Replacing Irrelevant Nouns}

\begin{table}[h!]
\centering
\small
\begin{tabular}{|l|c|c|}
\hline
\textbf{Models} & \textbf{Origin} & \textbf{Replace Nouns} \\
\hline
Qwen2.5 1.5B & 62.5\% & 54.9\% \\
Qwen2.5 7B & 83.5\% & 78.0\% \\
Qwen2.5 72B & 92.3\% & 81.9\% \\
Llama3.1 8B & 54.7\% & 51.7\% \\
Llama3.1 70B & 80.8\% & 74.2\% \\
GPT4o mini & 71.3\% & 64.1\% \\
GPT4o & 86.7\% & 79.5\% \\
\hline
\end{tabular}
\caption{Performance of replacing nouns on GSM8K. We report results on it since it has irrelevant nouns.}
\label{tab:}
\end{table}
\noindent\textbf{Replacing irrelevant nouns degrades performance consistently across various models.} As seen in Table 5, the scores of all models drop when the terms are renamed, with the magnitude of the decrease being similar across models. GPT4o models still show a decline.
\begin{table}[h!]
\centering
\small
\begin{tabular}{|c|c|c|c|c|}
\hline
\textbf{Models} & \textbf{Origin} & \textbf{High} & \textbf{Medium} & \textbf{Low} \\
\hline
Llama3.1 8B & 54.7\% & 51.5\% & 48.0\% & 44.0\% \\
Qwen2.5 7B & 83.5\% & 82.0\% & 78.1\% & 70.7\% \\
\hline
\end{tabular}
\caption{Model performance on replacing nouns with various semantic relevance levels.}
\end{table}

Replacing irrelevant nouns with semantically distant words further reduces the effectiveness.

\section{Discussion}~\label{discussion}
\subsection{Reasons Behind Accuracy Drops}
The above ablation of results reveals that LLMs are severely biased to common but irrelevant patterns. Now, we delve a little deeper into root causes.

\noindent\textbf{Could the imbalance in the test data be causing biased results?} On the MMLU benchmark, a naïve policy that always selects the longest option reaches 28.3\% accuracy, only +3.3\% above the 25 \% random baseline. This small gain shows that the length distribution of the options, by itself, is insufficient to yield the substantial performance gap reported in the paper.

\noindent\textbf{Could the failures of large language models be attributed to certain mechanisms, such as the attention mechanism?}
Yes, we perform an analysis of attention patterns and find that lengthening options affects the attention mechanisms, causing LLMs to attend more to that option.
\begin{table}[ht]
\centering
\small
\begin{tabular}{lcccc}
\hline
Condition & A    & B    & C    & D    \\
\hline
Origin    & 0.12 & 0.19 & \textbf{0.12} & 0.21 \\
WL        & 0.10 & 0.16 & \textbf{0.26} & 0.17 \\
\hline
\end{tabular}
\caption{The attention scores of options. The scores are from layer 0 and are averaged across all 32 heads over 24 option-orders; the numbers are the summed attention weights on the choice tokens. A is the correct answer, while C is an intentionally lengthened distractor.}
\label{tab:attention_scores}
\end{table}
 As in Table \ref{tab:attention_scores}, the “WL” row shows that increasing the length of option C shifts more attention toward it, confirming a length-induced bias in the attention mechanism.

\subsection{Generalization of Results}
\noindent\textbf{Could simple interventions (e.g., fine-tuning, CoT) address these vulnerabilities?} We investigate both aspects by adding simple mitigations (fine-tuning) and Chain-of-Thought (CoT) prompting. On MMLU with \texttt{Qwen2.5-7B}, fine-tuning on augmented perturbations narrows robustness variance: WL improves by $+7.8\%$ while RL decreases by $-12.3\%$ versus the base model, partially mitigating length bias. CoT further shrinks the RL--WL gap from $34.5\%$ to $15.4\%$, but lowers overall accuracy ($-10.7\%$), indicating that neither naive CoT nor simple fine-tuning fully resolves the vulnerability; stronger defenses (e.g., adversarial training or architectural changes) are likely required. CoT remains our default on GSM8K; here we additionally report MMLU results in Table~\ref{tab:mitigation_cot}.

\begin{table}[h]
\centering
\small
\begin{tabular}{lccc}
\toprule
Model & Origin & RL & WL \\
\midrule
Qwen2.5-7B & 73.7 & 90.1 & 55.6 \\
Qwen2.5-7B (fine-tuned) & 69.1 & 77.8 & 63.4 \\
Qwen2.5-7B (CoT) & 63.0 & 71.7 & 56.3 \\
\bottomrule
\end{tabular}
\caption{Accuracy (\%) on MMLU for \texttt{Qwen2.5-7B} under simple mitigations: \emph{fine-tuning} and \emph{Chain-of-Thought} (CoT).}
\label{tab:mitigation_cot}
\end{table}

\noindent\textbf{Could more proprietary LLMs, such as Gemini and Claude, be fragile to these changes?}

\begin{table}[t]
  \centering
  \small
  \begin{tabular}{lccc|cc}
    \toprule
    Model & Origin & RL & WL & MCQ & BQ \\
    \midrule
    Qwen2.5 72B        & 85.4 & 94.1 & 75.6 & 84.0 & 69.1 \\
    GPT4o              & 85.2 & 89.7 & 83.3 & 84.7 & 59.5 \\
    Deepseek-v3        & 86.4 & 90.4 & 84.3 & 85.0 & 71.2 \\
    Claude-3.5-sonnet  & 86.1 & 92.2 & 85.6 & 87.9 & 30.6 \\
    Gemini-2.0-flash   & 85.8 & 90.3 & 85.0 & 86.1 & 41.3 \\
    \bottomrule
  \end{tabular}
  \caption{Accuracy (\%) of SoTA short CoT models on MMLU.}
  \label{tab:shortcot-mmlu}
\end{table}
As we can see from Table~\ref{tab:shortcot-mmlu}, SoTA short CoT models, including Gemini-2.0-flash\footnote{Gemini 2: https://blog.google/technology/google-deepmind/google-gemini-ai-update-december-2024/}, Claude-3.5-sonnet\footnote{Claude 3.5 sonnet: https://www.anthropic.com/news/claude-3-5-sonnet}, and Deepseek-v3~\cite{deepseekai2025deepseekv3technicalreport}, are still sensitive to changes.
\subsection{Compare to Work in the Pre-LLM Era}
 Indeed, early studies have shown the limited robustness of LLMs from 2018 to 2020 \cite{naik-etal-2018-stress,zhang-etal-2019-paws, ribeiro-etal-2020-beyond}. However, those analyses were limited to less than 1B-parameter models, whereas modern LLMs exhibit strong generalization abilities that may alter robustness patterns. Our work extends this line by examining SoTa models on comprehensive and premium-quality benchmarks and uncovers significant limitations of these capable LLMs. 
% 文中引用：
% As we can see from Table~\ref{tab:shortcot-mmlu}, SoTA short CoT models, including
% Gemini-2.0-flash, Claude-3.5-sonnet, and Deepseek-v3, are still sensitive to changes
% in option length and problem type.

\subsection{Rank Changes of Models} As in \ref{sec:cg}, the rank of models on the MMLU leaderboard changes across different settings. 

\section{Conclusion}~\label{sec_conclusion}
This paper finds that LLMs exhibit significant performance degradation when faced with slight changes in question format, option length, or irrelevant content shifts. These findings underscore that LLMs rely on superficial patterns rather than robust, generalizable reasoning. By introducing the "Generalization Stress Tests," we offer novel understandings towards evaluating LLMs' true generalization capabilities. 
% This work aligns with the ACL 2025 theme on model generalization, advocating for developing more reliable benchmarks to assess LLMs beyond their superficial performance on traditional evaluation sets.

\section*{Limitations}~\label{sec_limitations}
This work focuses solely on non-chain-of-thought LLMs, such as GPT-4o, and does not consider emerging O1.

\section*{Ethnic Statement}~\label{sec_eth}
This work adheres to ACL's ethical guidelines, and we state that there are no ethical concerns to our knowledge.

\section*{Acknowledgments}
We thank the anonymous reviewers, the AC, SAC, and PC for their contributions to this work, and we are grateful to Change Jia for involvement in the early exploration and to Hongye Zhang for the manual annotation.
% 其实还有novely的问题， 这篇工作的相关工作很多

% \section*{Acknowledgments}

% This document has been adapted
% by Steven Bethard, Ryan Cotterell and Rui Yan
% from the instructions for earlier ACL and NAACL proceedings, including those for
% ACL 2019 by Douwe Kiela and Ivan Vuli\'{c},
% NAACL 2019 by Stephanie Lukin and Alla Roskovskaya,
% ACL 2018 by Shay Cohen, Kevin Gimpel, and Wei Lu,
% NAACL 2018 by Margaret Mitchell and Stephanie Lukin,
% Bib\TeX{} suggestions for (NA)ACL 2017/2018 from Jason Eisner,
% ACL 2017 by Dan Gildea and Min-Yen Kan,
% NAACL 2017 by Margaret Mitchell,
% ACL 2012 by Maggie Li and Michael White,
% ACL 2010 by Jing-Shin Chang and Philipp Koehn,
% ACL 2008 by Johanna D. Moore, Simone Teufel, James Allan, and Sadaoki Furui,
% ACL 2005 by Hwee Tou Ng and Kemal Oflazer,
% ACL 2002 by Eugene Charniak and Dekang Lin,
% and earlier ACL and EACL formats written by several people, including
% John Chen, Henry S. Thompson and Donald Walker.
% Additional elements were taken from the formatting instructions of the \emph{International Joint Conference on Artificial Intelligence} and the \emph{Conference on Computer Vision and Pattern Recognition}.

\bibliography{acl_latex}

\appendix

\section{Prompts and Verification in Altering Option Length}
\label{sec:appendix-length}
\subsection{Prompts}
We chose the GPT-4o to lengthen options.

\textbf{The default prompt to lengthen options is:} The user will give you a question, the choices, and the answer from a dataset. Rewrite the four choices into longer ones. Make sure not to change the question willingly. Make sure that the rewritten options do not contain a hint of the correct answer.

\textbf{The prompt to control option length is:}
We concatenate the default prompt to one of the following prompts.
\begin{itemize}
    \item Make sure that each rewritten option contains no more than 10 words.
    \item Make sure that each rewritten option at least 10 words and no more than 20 words.
    \item Make sure that each rewritten option contains at least 20 words.
\end{itemize}

We set the temperature to $0$, and the other setting is the same as the default.

\subsection{Verification Process}
We manually verified the rewritten sentences to check whether lengthening the sentence introduced factors related to the answer or changed the question's meaning. We manually checked 100 examples from MMLU and found that 99 had no issues, while 1 changed the original meaning of the question. The rewriting accuracy was 99\%.
\section{Prompts in Replacing Irrelevant Nouns}
\label{sec:appendix-noun}
% \subsection{Generation Details}
We found that GPT-4o and GPT-4o mini perform similarly on this task. To reduce carbon emissions, we chose the GPT-4o mini. 

\textbf{The prompt to simply replace irrelevant nouns is:} Assist in creatively substituting nouns in mathematical problems to prevent students from memorizing solutions. The replacements should be imaginative, ensuring the mathematical relationships and the accuracy of the solutions are preserved. ``{input\_text}'' Other than replacing nouns, do not alter the original word order sentence structure, or add or remove any sentences. Give the modified question directly.

\textbf{The prompt to alter semantic relevance is: }
Substitute nouns and some relevant words in the mathematical problems creatively to prevent students from memorizing solutions. The replacements should be done in three levels: 

\begin{itemize}
    \item Level 1: Only replace nouns with semantically similar words (e.g., 'apple' becomes 'banana').
    \item Level 2: Replace nouns and verbs with words that differ in meaning but are still within the realm of common sense (e.g., 'apple' becomes 'elephant', 'eat fruit' becomes 'drink coke').
    \item Level 3: Replace words as much as possible with highly imaginative and fantastical words, if you think it still makes sense in mathematical problems. (e.g., 'apple' becomes 'alien gemstone'). 
\end{itemize}
Apart from replacing nouns and some relevant words, maintain the original word order, sentence structure, and do not add or remove any sentences. Give three modified sentences directly, one for each level, only separated by '\#\#\#'. Don't return anything else including 'Level 1', 'Level 2', 'Level 3' but only "\#\#\#". This is the original question: {input\_text}     

We set temperature to $0.1$, top-p to $1$, top-k to $0$, and repetition\_penalty to $0$.

% \subsection{Verification Process}

\section{Experiment Setup Details}
\label{sec:appendix-exp}
This section describes the foundational setup of our experiments and analyses, including the evaluation framework and methods we used and the benchmarks and models we evaluated.
\begin{table}
  \centering
  \small
  \begin{tabular}{llllll}
    \hline
    \textbf{Benchmark} & \textbf{Model}& \textbf{Origin}& \textbf{AL}  & \textbf{RL} & \textbf{WL} \\
    \hline
    \multirow{8}{*}{MMLU} & Qwen2.5 1.5B & \textbf{60.3} & \textbf{54.7} & \textbf{89.0} & \textbf{36.3}\\
                           & Qwen2.5 7B & 73.7&69.2 &90.1  &55.6 \\
                           & Qwen2.5 72B & 85.4&81.3 & 94.1 & 75.6 \\
                           % &Qwen2 1.5B &55.8 &90.0 &27.3\\
                           % & Qwen2 7B & 70.6 & 87.8 & 57.5 \\
                           % & Qwen2 72B & 83.8 & 94.1 & 74.1 \\
                           & LLaMa3.1 8B & 65.5&64.3 & 85.6 & 53.6 \\
                           & LLaMa3.1 70B & 78.8 &76.0 & 93.6 & 70.6 \\
    \hline
    % \multirow{8}{*}{ARC-C} & Qwen2 1.5B & 69.0 & 88.3 & 53.8 \\
    %                        & Qwen2 7B & 87.2 & 93.3 & 83.8 \\
    %                        & Qwen2 72B & 95.7 & 97.3 & 93.1 \\
    \multirow{8}{*}{ARC-C} & Qwen2.5 1.5B & 77.3&67.3 & 88.9 & 68.1 \\
                           & Qwen2.5 7B & 90.0&85.3 & 94.3 & 84.0 \\
                           & Qwen2.5 72B & 95.8&93.1 & 97.2 & 94.4 \\
                           & LLaMa3.1 8B & 78.1&78.6 & 85.2 & 74.7 \\
                           & LLaMa3.1 70B & 91.8&89.9 & 96.3 & 90.8 \\
    \hline
  \end{tabular}
  \caption{\label{fig:AL}
    Performance on altering option length. AL refers to lengthening all options. RL refers to lengthening the right option. WL refers to lengthening the wrong option. The values are percentages.
  }
\end{table}
\subsection{Evaluation Protocol}
We perform evaluations on the Harness framework \cite{eval-harness}. We chose Harness because it is a flexible, configurable, reproducible framework. Unless specified, we follow the default parameter of the harness.
Unless otherwise specified, our evaluations are conducted in a 5-shot manner, with few-shot examples drawn from the benchmarks' corresponding training sets. 
\subsection{Models}
We evaluate models of Llama3.1 series \cite{llama31}, Qwen2 series \cite{qwen2}, and GPT4o. Llama3.1 and Qwen2.5  are the most powerful small models, while GPT4o is the most powerful LLM. We list all models below.
\begin{itemize}
    \item Llama3.1 8B, Llama3.1 70B;
    % \item Gemma2 2B, Gemma2 9B, Gemma2 27B;
    \item Qwen2.5 1.5B, Qwen2.5 7B, Qwen2.5 72B;
    \item GPT4o, GPT4o mini.
\end{itemize}
\subsection{Benchmarks}
% 改写命令，gsm8k的多选题
We evaluate LLMs on MMLU, ARC, Helaswag, GSM-MCQ, and GSM8k. The first four are MCQ benchmarks, and the last consists of open-ended questions.
\begin{itemize}
\item \textbf{MMLU}~\cite{mmlu} is a multi-task benchmark that covers 57 tasks ranging from elementary to college level. These tasks cover multiple disciplines, e.g., math, physics, law, history, etc. The whole test set consists of 14,042 examples.  Following common practice, we calculate the accuracy of each task and report the average score across all tasks.
\item \textbf{ARC}~\cite{arc} is also a multitask dataset that includes data from eight types of tasks, testing aspects such as common sense, multi-hop reasoning, and algebraic operations, with 3,548 samples. ARC has two subsets: one is ARC-Challenge (abbreviated as ARC-C), and the other is ARC-Easy (abbreviated as ARC-E). The challenge set includes only those data that cannot be answered through retrieval and word co-occurrence methods, making it more difficult.
% \item \textbf{Hellaswag}~\cite{hellaswag} HellaSwag is a benchmark that tests whether a model can complete sentences. It is also a multiple-choice question benchmark, where the questions consist of incomplete sentences, and the options are several potential completions. This benchmark contains a total of 10,003 test samples.
\item \textbf{GSM-8K}~\cite{gsm} examines multi-step math word problems, which are relatively easy and designed to be solvable by middle school students. GSM8K is presented in an open-ended question format, unlike multiple-choice questions. It consists of 1,319 test questions.
% \item \textbf{GSM-MCQ}~\cite{zhang2024multiplechoicequestionsefficientrobust} is a MCQ version of GSM-8k. The wrong options are incorrect predictions from 60 open-source LLMs.
\end{itemize}
\subsection{Budget}
We performed experiments with an H800 GPU; the total cost of the experiments was about 1000 GPU hours.

\subsection{Random Seeds}
% Required packages: \usepackage{multirow}
All reported numbers are from a single wrong, since the model is deterministic with our default decoding temperature \(T=0\), thus, changing seeds has no effect. We nevertheless tested robustness at \(T=0.5\) and \(T=1.0\) while varying Python/NumPy/Torch seeds; results for Qwen-2.5 7B on MMLU are shown below. The RL/WL gap persists, and our conclusions remain unchanged.

\begin{table}[h!]
\centering
\setlength{\tabcolsep}{7pt}
\renewcommand{\arraystretch}{1.15}
\begin{tabular}{c l c c c}
\hline
\textbf{Temp.} & \textbf{Seed} & \textbf{Origin} & \textbf{RL} & \textbf{WL} \\
\hline
\multirow{3}{*}{0}   & 0, 1234, 1234 & 73.7 & 90.1 & 55.7 \\
                     & 1, 11, 111    & 73.7 & 90.1 & 55.7 \\
                     & 2, 22, 222    & 73.7 & 90.1 & 55.7 \\
\hline
\multirow{3}{*}{0.5} & 0, 1234, 1234 & 71.6 & 88.2 & 54.9 \\
                     & 1, 11, 111    & 71.5 & 88.5 & 54.9 \\
                     & 2, 22, 222    & 71.5 & 88.3 & 55.3 \\
\hline
\multirow{3}{*}{1.0} & 0, 1234, 1234 & 67.9 & 83.4 & 52.2 \\
                     & 1, 11, 111    & 67.3 & 83.8 & 51.9 \\
                     & 2, 22, 222    & 67.2 & 83.8 & 53.0 \\
\hline
\end{tabular}
\caption{Model accuracy (\%) for Qwen-2.5 7B on MMLU under different temperatures and seeds. Numbers are averaged over three runs when applicable; at \(T=0\) the model is deterministic. Each seed setting contains three seeds regarding Python, NumPy, and Torch.}
\label{tab:seed_temp_mmlu}
\end{table}

\section{Additional Results}
\subsection{Making All Options longer}
\label{sec:al}

We can see from Table~\ref{fig:AL} that LLaMa is more robust than Qwen, and larger models are more robust than smaller models, when we make all options longer. Besides, even if we introduce the setting of AL, our conclusion that LLMs are vulnerable to option lengths and biased to long options is not changed. 
\subsection{Make All Wrong Options Longer}
\label{sec:wl_all}
\begin{table}[ht]
\centering
\begin{tabular}{|l|c|c|c|}
\hline
\textbf{Model} & \textbf{origin} & \textbf{WL} & \textbf{WL-ALL} \\
\hline
Llama3.1 8B & 65.5\% & 53.6\% & 64.8\% \\
Llama3.1 70B & 78.8\% & 70.6\% & 82.4\% \\
gpt-4o & 85.2\% & 83.3\% & 85.6\% \\
\hline
\end{tabular}
\caption{Results of making all wrong options longer on the MMLU benchmark.}
\label{tab:wl_all}
\end{table}

\noindent\textbf{Making all wrong options could expose the right answer.} From Table \ref{tab:wl_all}, we can see that if all the incorrect options are lengthened, the model will choose the only correct option that hasn't been lengthened.
\subsection{Rank Changes of Models}
\label{sec:cg}
% Requires: \usepackage{booktabs}
% Optional: \usepackage{siunitx} for number alignment

\begin{table*}[h]
\centering
\caption{Accuracy (\%), rank, and rank changes of models on MMLU under different protocols.}
\label{tab:mmlu-rl-wl}
\begin{tabular}{lrrrrrrrr}
\toprule
\multirow{2}{*}{Model} & \multicolumn{2}{c}{Origin} & \multicolumn{3}{c}{RL} & \multicolumn{3}{c}{WL} \\
\cmidrule(lr){2-3}\cmidrule(lr){4-6}\cmidrule(lr){7-9}
 & Score & Rank & Score & Rank & $\Delta$Rank & Score & Rank & $\Delta$Rank \\
\midrule
Qwen2.5 1.5B & 60.3 & 7 & 89.0 & 5 & $\uparrow\,2$ & 36.3 & 7 & --- \\
Qwen2.5 7B   & 73.7 & 5 & 90.1 & 3 & $\uparrow\,2$ & 55.6 & 5 & --- \\
Qwen2.5 72B  & 85.4 & 1 & 94.1 & 1 & ---        & 75.6 & 2 & $\downarrow\,1$ \\
LLaMa 3.1 8B & 65.5 & 6 & 85.6 & 7 & $\downarrow\,1$ & 53.6 & 6 & --- \\
LLaMa 3.1 70B& 78.8 & 3 & 93.6 & 2 & $\uparrow\,1$ & 70.6 & 3 & --- \\
GPT\mbox{-}4o mini & 76.5 & 4 & 87.2 & 6 & $\downarrow\,2$ & 70.6 & 3 & $\uparrow\,1$ \\
GPT\mbox{-}4o      & 85.2 & 2 & 89.7 & 4 & $\downarrow\,2$ & 83.3 & 1 & $\uparrow\,1$ \\
\midrule
Kendall $\tau$ & \multicolumn{2}{c}{---} & \multicolumn{3}{c}{0.52} & \multicolumn{3}{c}{0.88} \\
\bottomrule
\end{tabular}
\end{table*}

\begin{table}[h]
\centering
\caption{Accuracy (\%), rank, and rank changes of models on MMLU (MCQ vs.\ BQ).}
\label{tab:mmlu-mcq-bq}
\begin{tabular}{lrrrrr}
\toprule
\multirow{2}{*}{Model} & \multicolumn{2}{c}{MCQ} & \multicolumn{3}{c}{BQ} \\
\cmidrule(lr){2-3}\cmidrule(lr){4-6}
 & Score & Rank & Score & Rank & $\Delta$Rank \\
\midrule
Qwen2.5 1.5B & 58.8 & 7 & 30.3 & 7 & --- \\
Qwen2.5 7B   & 72.4 & 5 & 54.7 & 4 & $\uparrow\,1$ \\
Qwen2.5 72B  & 84.0 & 2 & 69.1 & 1 & $\uparrow\,1$ \\
LLaMa 3.1 8B & 64.6 & 6 & 40.6 & 6 & --- \\
LLaMa 3.1 70B& 78.4 & 3 & 63.5 & 2 & $\uparrow\,1$ \\
GPT\mbox{-}4o mini & 75.1 & 4 & 54.5 & 5 & $\downarrow\,1$ \\
GPT\mbox{-}4o      & 84.7 & 1 & 59.5 & 3 & $\downarrow\,2$ \\
\midrule
Kendall $\tau$ & \multicolumn{2}{c}{---} & \multicolumn{3}{c}{0.71} \\
\bottomrule
\end{tabular}
\end{table}

\noindent\textbf{Note of Kendall $\tau$ .} Kendall $\tau$ is a non-parametric rank-correlation coefficient that measures how similarly two lists are ordered by comparing the balance of concordant versus discordant item pairs, ranging from $-1$ (complete disagreement) to $+1$ (perfect agreement).

\noindent As we can see, the rank of the LLMs changes when we apply the perturbation. The Kendall $\tau$ between the WL and RL rankings is 0.39; since a Kendall $\tau$ below 0.4 is not considered strongly correlated, \textbf{switching the evaluation protocol still has a pronounced influence on the rank shifts}.

% \subsection{Make ALL Wrong Options Longer.}
% We further verify whether large models possess invariant knowledge across changes in question types.

\end{document}